\documentclass{article}
\usepackage{adjustbox}

\usepackage[preprint]{neurips_2025}
\usepackage[utf8]{inputenc} 
\usepackage[T1]{fontenc}    
\usepackage{hyperref}       
\usepackage{url}            
\usepackage{booktabs}       
\usepackage{enumitem}      
\usepackage{amsfonts}       
\usepackage{nicefrac}       
\usepackage{microtype}      
\usepackage{comment}
\usepackage{textcomp}
\usepackage{amsmath}
\usepackage{bm}
\usepackage{multirow} 
\usepackage{siunitx}  
\usepackage{graphicx} 
\usepackage{threeparttable}
\usepackage{colortbl}

\usepackage{algorithm}
\usepackage{algpseudocode}  
\usepackage{multirow}
\usepackage{pifont}
\usepackage[most]{tcolorbox}  
\usepackage{wrapfig}
\usepackage{tikz}
\usepackage{graphicx} 
\usepackage{makecell}
\usepackage{epigraph}
\usepackage{caption}
\usepackage[framemethod=TikZ]{mdframed} 

\definecolor{myblue}{rgb}{0.0, 0.25, 1.0}
\definecolor{lightblue}{RGB}{194, 223, 255}
\definecolor{mygreen}{RGB}{35, 153, 78}
\newcommand{\deltavalue}[2]{#1\scalebox{0.7}{\textcolor{gray}{{ + #2}}}}
\newcommand{\oursvalue}[2]{#1\scalebox{0.7}{\textbf{\textcolor{myblue}{{ + #2}}}}}
\newcommand{\negvalue}[2]{#1\scalebox{0.7}{\textcolor{mygreen}{{ - #2}}}}

\usepackage{listings}

\usepackage{xcolor} 

\usepackage{lipsum}           

\definecolor{light-gray}{gray}{0.95} 

\newmdenv[
  linecolor=gray!40, 
  outerlinewidth=0.5pt, 
  roundcorner=5pt, 
  innertopmargin=10pt, 
  innerbottommargin=10pt, 
  innerrightmargin=10pt, 
  innerleftmargin=10pt, 
  backgroundcolor=gray!5, 
  skipabove=\baselineskip, 
  skipbelow=\baselineskip 
]{elegantbox}

\newtcolorbox{quotebox}{
    colback=gray!00, 
    colframe=gray!00, 
    coltext=black,
    boxrule=0pt,
    arc=4pt,
    leftmargin=10pt,
    right=10pt,
    top=1pt,
    bottom=1pt,
}

\title{Generative RLHF-V: Learning Principles from Multi-modal Human Preference}

\author{%
\textbf{Jiayi Zhou\thanks{Equal contribution, $^{\dag}$Corresponding author.}\,~$^{,1}$,} \textbf{Jiaming Ji$^{*, 1}$,} \textbf{Boyuan Chen$^{1}$,} \textbf{Jiapeng Sun$^{3}$,} \textbf{Wenqi Chen$^{1}$} \\ \textbf{Donghai Hong$^{1}$,} \textbf{Sirui Han$^{2}$,} \textbf{Yike Guo$^{2}$,} \textbf{Yaodong Yang}$^{\dag, 1}$ 
\\
\vspace{-0.5em}
\\
$^{1}$Peking University
\\ 
$^{2}$Hong Kong University of Science and Technology
\\ 
$^{3}$University College London
\\
\texttt{\{gaiejj,jiamg.ji,cbylll\}@stu.pku.edu.cn}\\
\texttt{yaodong.yang@pku.edu.cn} \vspace{-0.5em}
}
\begin{document}

\maketitle

\begin{abstract}

Training multi-modal large language models (MLLMs) that align with human intentions is a long-term challenge. Traditional score-only reward models for alignment suffer from low accuracy, weak generalization, and poor interpretability, blocking the progress of alignment methods, \textit{e.g.,} reinforcement learning from human feedback (RLHF). Generative reward models (GRMs) leverage MLLMs' intrinsic reasoning capabilities to discriminate pair-wise responses, but their pair-wise paradigm makes it hard to generalize to learnable rewards. We introduce Generative RLHF-V, a novel alignment framework that integrates GRMs with multi-modal RLHF. We propose a two-stage pipeline: \textbf{multi-modal generative reward modeling from RL}, where RL guides GRMs to actively capture human intention, then predict the correct pair-wise scores; and \textbf{RL optimization from grouped comparison}, which enhances multi-modal RL scoring precision by grouped responses comparison. Experimental results demonstrate that, besides out-of-distribution generalization of RM discrimination, our framework improves 4 MLLMs' performance across 7 benchmarks by 18.1\%, while the baseline RLHF is only 5.3\%. We further validate that Generative RLHF-V achieves a near-linear improvement with an increasing number of candidate responses. Our code and models can be found at \url{https://generative-rlhf-v.github.io}.

\end{abstract}

\vspace{-0.5em}
\begin{figure}[htbp]
  \centering
  \includegraphics[width=\textwidth]{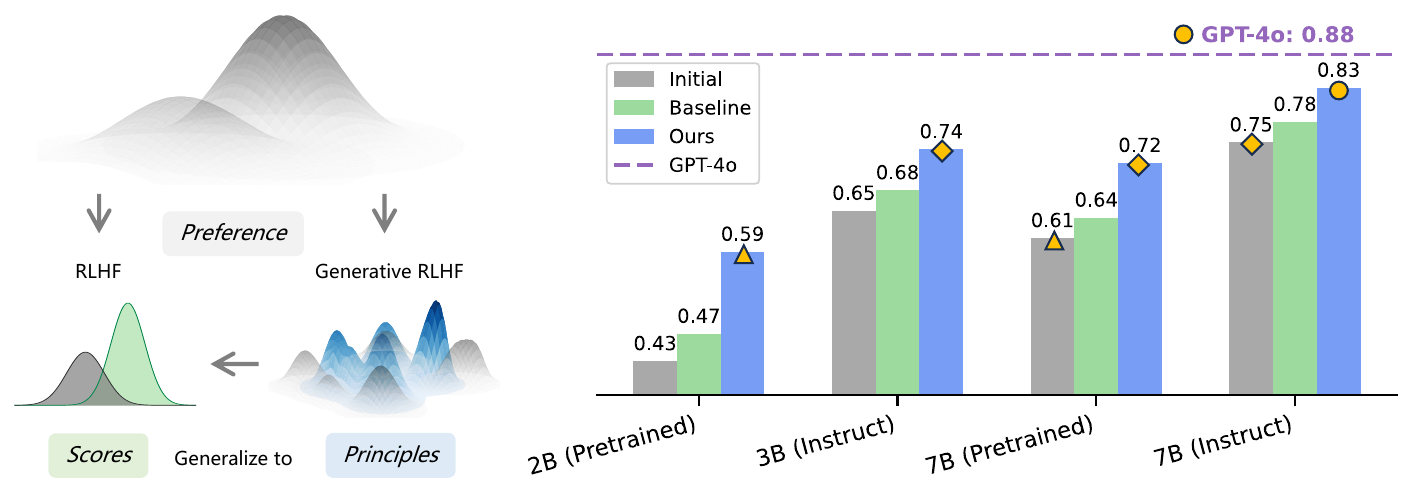}
  \vspace{-1.0em}
  \caption{Advanced multi-modal large language models (MLLMs) is calling principled preference learning. In MLLM's alignment, traditional RLHF methods only learn scalar scores from preferences. In contrast, our Generative RLHF-V can learn principles from preferences and optimize based on this comprehensive comparison. Experimental results show that Generative RLHF-V elevates 2B and 3B MLLMs to 7B performance across 7 benchmarks. It also advances pretrained models to instruct model capabilities and enables open-source models to match closed-source experts.}
  \vspace{-1.0em}
\end{figure}

\section{Introduction}
\label{sec:intro}

\epigraph{
    \textbf{"The mediocre teacher tells. The great teacher inspires."}}{
    --- William Arthur Ward saying – Education
}

 Human interaction and learning are naturally multi-modal \citep{rahwan2019machine, pmlr-v235-huh24a}. Recent research has demonstrated significant advances in multi-modal large language models (MLLMs) \citep{anil2023palm2, yao2024minicpm, team2024gemini, zhu2025internvl3} on visual question answering and reasoning tasks. These breakthroughs reveal two critical insights: 1) Reinforcement learning (RL) substantially enhances MLLMs' capacity for solving complex problems \citep{liu2025visual}; 2) The efficacy of RL fundamentally depends on the precisely defined reward (\textit{e.g.}, rule-based verification for mathematical correctness). While rule-based rewards can be effectively constructed for logical reasoning and factual judgment tasks \citep{guo2025deepseek, shen2025vlm}, accurate reward modeling for human values, \textit{e.g.,} instruction-following or safety \citep{ouyang2022training, bai2022training}, remains a long-term challenge in MLLMs alignment.

Alignment aims to make AI systems adhere to human intentions \citep{ji2023ai,bai2022training}, and for MLLMs, these goals can be concretized into the 3H standards: helpful, harmless, and honest \citep{askell2021general-3H, yu2024rlhf, ding2025mm}. These goals are difficult to represent as a symbolic reward \citep{openai2023gpt4, dai2024safe}. Traditional approaches typically employ an additional score head to project the final-layer activations of MLLMs into scalar rewards \citep{ouyang2022training, guo2025deepseek, sun2024aligning}, \textit{i.e.,} score-only reward model (RM), subsequently applying Bradley-Terry loss to learn human preferences from pairwise comparisons. However, extensive studies \citep{gao2023scaling, zhou2025sequence, ankner2024critique} have exposed three fundamental limitations of this paradigm: low accuracy, weak generalization, and poor interpretability. Generative reward models (GRMs) \citep{zhang2generative, mahan2024generative} present a promising alternative by leveraging LLMs' intrinsic reasoning capabilities to discriminate pair-wise responses \citep{wang2025unified, xiong2024llava}, and rule-based RL fine-tuning strengthens this capability \citep{liu2025inference}. Nevertheless, the practical application of such GRMs in multi-modal RLHF remains to be verified.  This progression presents the following urgent dilemma:

\begin{itemize}[leftmargin=0.3em]
    \item \textbf{Advanced MLLMs is calling principled preference learning.} As MLLMs become more sophisticated, they handle increasingly complex inputs and diverse tasks \citep{liang2024survey}. Human assessment of preferences for MLLMs' responses will also grow more varied and intricate \citep{ji2025survey}. Consequently, relying on a single inference from a score-only RM proves insufficient for learning generalizable human preferences \citep{ye2024beyond, gao2023scaling}, thereby creating a bottleneck in MLLMs alignment.
    \item \textbf{Pair-wise comparison is blocking multi-modal principles from generalizing to learnable rewards.} While pair-wise comparison allows GRM to learn generalizable principles from RL \citep{liu2025inference, zheng2023judging, chen2024mllm}, this pair-wise comparison feedback does not readily translate into the point-wise scores, which are essential for RL optimization \citep{xu2025unified, wu2023pairwise, hu2025open}. This disconnect hinders the ability of learned principles to effectively guide the multi-modal RLHF.
\end{itemize}


In response, we propose \textbf{Generative RLHF-V}(ision), as shown in \autoref{fig:main}, a novel alignment framework enabling the pair-wise multi-modal GRM with multi-modal RLHF. Our pipeline consists of two stages: \textbf{multi-modal generative reward modeling from RL} and \textbf{RL optimization from grouped comparison}. The first component utilizes RL to train a GRM to learn principles from multimodal preferences, which then performs strongly generalizable pairwise scoring of responses. The second component applies these GRM-learned principles to obtain more precise scores by comparing within groups of responses. Our GRM training extends the self-principled critique tuning (SPCT) \cite{liu2025inference} to the vision scenario, training MLLMs as GRMs using RL, with rule-based rewards from annotated ground truth in preference datasets. 

In contrast to SPCT, we find that in the multi-modal scenario, enabling GRMs to explore principles autonomously yields superior generalization than selecting principles from a reference set. Our grouped comparison design enables the generalization of learned principles from pair-wise comparisons to point-wise scores. This further unveils a novel direction for post-training scaling up: as the number of candidate responses $n$ explored by online RL increases, GRMs can assign more accurate scores, leading to improved RL performance near linearly. Our key contributions are as follows: 

\begin{figure}[t]
  \centering
  \includegraphics[width=\textwidth]{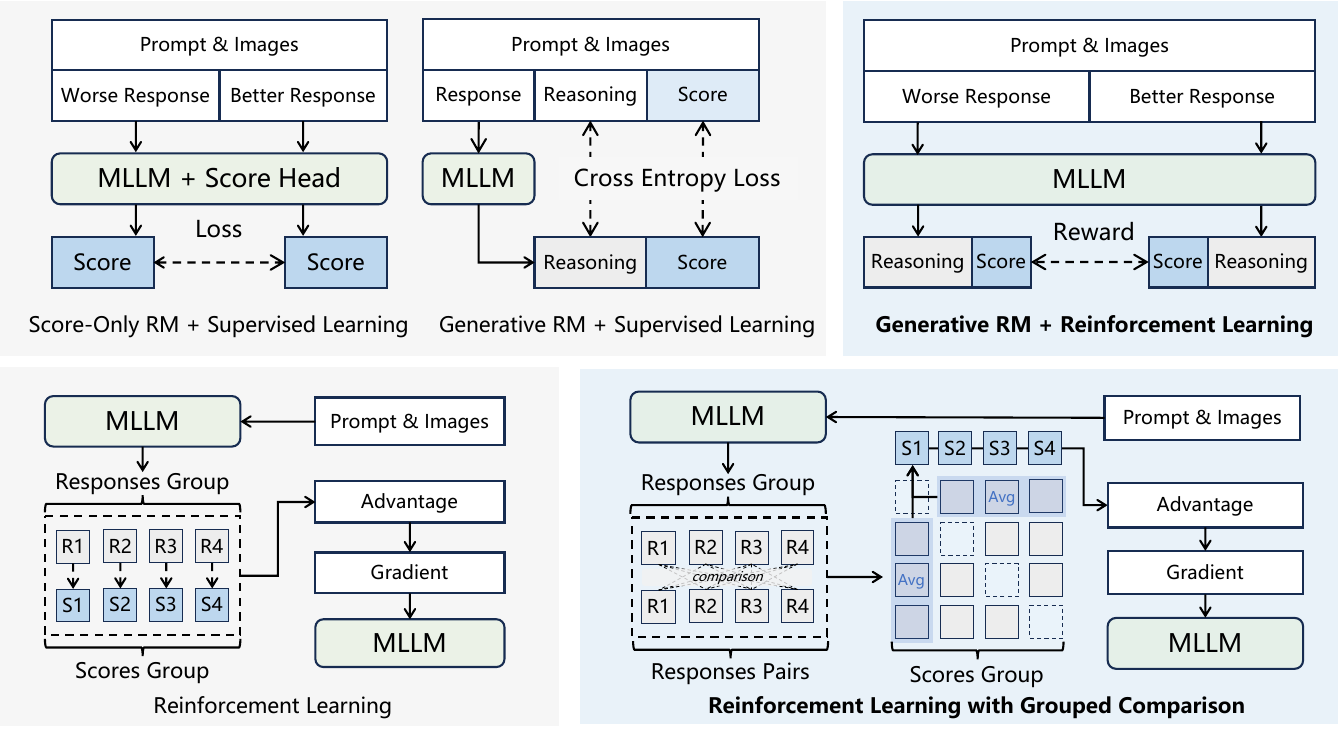}
  \vspace{-1.0em}
  \caption{Comparison of our pipelines to traditional ones. \textbf{For reward modeling}, we make generative RM actively reason about the advantages and disadvantages between two answers, and output corresponding scores. If the better response gets a higher score, it provides a positive reward. \textbf{For RL optimization}, we compare responses in pairs within a group to obtain more accurate scores.}
  \label{fig:main}
\end{figure}

\begin{itemize}[leftmargin=0.3em]
    \item \textbf{RL-based GRMs for learning principles from multi-modal preference:} We develop a multi-modal GRM trained via RL, enabling the reasoning of principles and precise reward predictions, achieving average \textbf{20.4\%} accuracy improvement on out-of-distribution discriminative tasks.
    \item \textbf{Multi-modal generative RLHF:} We empirically demonstrate the superiority of GRMs for multi-modal RLHF. Experimental results demonstrate that our framework improves MLLMs’ performance across 7 benchmarks by \textbf{18.1\%}, while the baseline RLHF is only \textbf{5.3\%}.  
    \item \textbf{Grouped comparison for post-training scaling up:} We discovered that the integration of GRM+RL with grouped comparison enables the performance of RL optimization to near linearly improve with an increasing number of candidate responses $n$ within a certain range. The removal of either component negates this observed enhancement.  
    \item \textbf{A pioneer case study of multi-modal GRM reward hacking:} We find that RL over-training under an over-trained GRM can lead models to adopt \textit{self-praise} behaviors to obtain high rewards, even achieving exceptionally high scores on benchmarks employing the MLLM-as-judge paradigm.
\end{itemize}

\section{Related Work and Preliminaries}

\textbf{MLLM Alignment and RLHF.} AI alignment is the deliberate process of shaping model behavior to cohere with human goals, values, and ethical principles \citep{ouyang2022training, ji2023ai, bai2022constitutional, bai2022training, rafailov2023direct, ke2023critiquellm}. Achieving robust alignment faces challenges in translating complex, subjective, and evolving human values into quantifiable training objectives \citep{gao2023scaling, zang2025internlm}. Current MLLM alignment methods mainly rely on post-training \citep{yu2024rlaif, zhang2025mm, sun2024aligning}, with RL fine-tuning based on human preferences being the most mainstream approach. This process typically involves two key stages: reward modeling and RL optimization.

A score-only reward model $R_\theta$ is trained on a dataset of human preferences, where each data point includes a prompt $\bm{x}$, a preferred response $\bm{y}^w$, and a dispreferred response $\bm{y}^l$. The model learns to assign a higher scalar score to the preferred response $s^w$ than the dispreferred one $s^l$ using a pairwise ranking loss, typically minimizing $\mathcal{L}_{RM} = - \sum \log \sigma(s^w_i - s^l_i)$. This trained $R_\theta$ serves as an automated judge of response quality.  The loss function can be expressed as minimizing the negative log-likelihood over the dataset:
$$\mathcal{L}_{RM}(\theta) = - \sum_{i=1}^{N} \log \sigma(R_\theta(\bm{x}_i, \bm{y}^w_i) - R_\theta(\bm{x}_i, \bm{y}^l_i)) = - \sum_{i=1}^{N} \log \sigma(s^w_i - s^l_i).$$
The MLLM's policy ($\pi_{\phi}$) is fine-tuned using RL. For a given prompt $\bm{x}$ from a given dataset $\mathcal{D}_{\text{prompt}}$, the policy generates a response $\bm{y}$, which is then scored by the reward model $R_\theta$. The policy parameters $\phi$ are updated to maximize this reward. To prevent the policy from deviating too much from the original pre-trained model ($\pi_{\phi}^{\text{base}}$) and maintain coherence, a Kullback-Leibler (KL) divergence penalty is added to the optimization objective: $\max_{\phi} \mathbb{E} [R_\theta(\bm{x}, \bm{y}) - \beta \text{KL}(\pi_{\phi} || \pi_{\phi}^{\text{base}})]$, where $\beta$ is a fixed hyper-parameter. The final optimization objective is:
$$\max_{\phi} \mathbb{E}_{\bm{x} \sim \mathcal{D}_{\text{prompt}}, \bm{y} \sim \pi_{\phi}(\cdot|\bm{x})} [R_\theta(\bm{x}, \bm{y}) - \beta \text{KL}(\pi_{\phi}(\cdot|\bm{x}) || \pi_{\phi}^{\text{base}}(\cdot|\bm{x}))].$$
\textbf{Generative Reward Model.} Generative reward models (GRM) \citep{mahan2024generative, zhang2generative, ankner2024critique} offer an alternative paradigm to score-only reward modeling, which leverages the inherent generative capabilities of MLLMs to evaluate preferences. Current research on GRMs for MLLM alignment focuses on employing supervised learning methods to improve accuracy \citep{wang2025unified}. A representative method is LLaVA-Critic \citep{xiong2024llava}, which collects expert-annotated point-wise and pair-wise scores, along with reasoning traces for MLLM question-answer pairs, subsequently training the MLLM as a GRM via supervised learning. Despite its superior performance, this approach necessitates more expensive expert annotations compared to binary preference datasets and imposes stricter requirements for the reasoning trace annotations. Moreover, there is a notable lack of empirical studies on applying GRMs in RL training. To date, explorations of GRM applications have centered on data filtering for Best-of-N selection and offline direct preference optimization \citep{rafailov2023direct}. The practical implementation of GRM within the multi-modal RL optimization is yet to be investigated.





\section{Generative RLHF-V}

The Generative RLHF-V pipeline mainly consists of two parts: generative reward modeling from reinforcement learning (RL) and RL from grouped comparison. The former references training MLLMs through RL as a pair-wise vision generative reward model (GRM), which actively reasons about the human principle behind two given responses and provides a pair-wise score comparison. The latter leverages the characteristics of this GRM, collecting multiple responses for a given input and providing more accurate grouped scoring for them.

\begin{figure}[t]
  \centering
  \includegraphics[width=\textwidth]{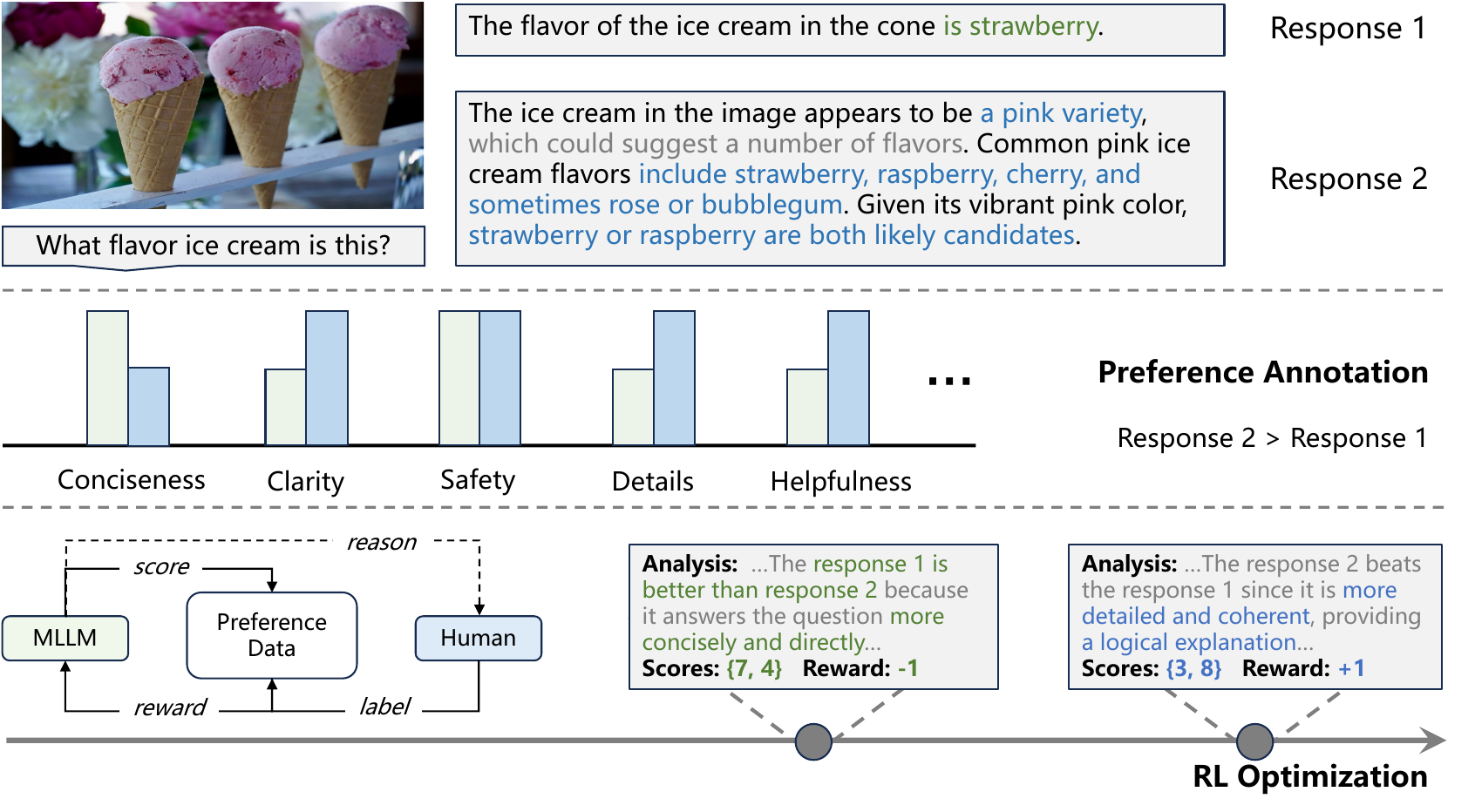}
  \vspace{-1.3em}
  \caption{An example of generative reward modeling from RL. The goal of RL is to make MLLMs assign higher scores to responses that align with human preferences. Through RL optimization, MLLMs can infer the underlying principle behind how humans annotate these binary preferences.}
\end{figure}

\textbf{Multi-modal Generative Reward Modeling from RL.} We consider the task of training a pair-wise GRM $R_\theta$ with parameters $\theta$ using RL guided by human preferences. The goal of $R_\theta$ is to assign a pair of scalar scores $\{\bm{s}_1, \bm{s}_2\}$ to a pair of responses $\{\bm{y}_1, \bm{y}_2\}$ for a given prompt $\bm{x}$. We are given a dataset of human preferences $\mathbb{D}^{\mathbb{P}} = \{(\bm{x}_i, \bm{y}^w_i, \bm{y}^l_i)\}_{i=1}^N$, where $\bm{x}_i$ is a prompt, $\bm{y}^w_i$ is the response preferred and $\bm{y}^l_i$ is the response dispreferred by a set of human preference principles $\mathbb{P}=\{p_1, p_2, \dots, p_k\}$. The inference of reward model, denoted as $\pi^{\theta}_{\text{GRM}}$ (or simply $R_\theta$), is a parameterized function that takes the prompt $\bm{x}$, and responses pairs $\bm{y}^w$ and $\bm{y}^l$ as input and outputs the predicted principles $\mathbb{P}^*$, reasoning traces $r$, and a pair-wise scalar score $s^w$ and $s^l$:
$$\{\mathbb{P}^*, r, s^w, s^l\} = R_\theta(\bm{x}, \bm{y}^w, \bm{y}^l).$$
The model's preference prediction should be $s_w > s_l$. And the reward $r$ for a given preference pair $(\bm{x}, \bm{y}^w, \bm{y}^l)$ is determined by comparing the scores assigned by $R_\theta$:
$$r(\bm{x}, \bm{y}^w, \bm{y}^l; \theta) = \begin{cases} +1 & \text{if } s^w > s^l, \\ -1 & \text{if } s^w \le s^l. \end{cases}$$
The RL objective is to maximize the expected reward over the preference dataset $\mathcal{D}$.
$$\max_{\theta} \mathbb{E}_{(\bm{x}, \bm{y}^w, \bm{y}^l) \sim \mathcal{D}} \left[ r(\bm{x}, \bm{y}^w, \bm{y}^l; \theta) \right].$$

\textbf{Reinforcement Learning from Grouped Comparison.} This stage is to fine-tune the MLLM, denoted as $\pi_{\phi}$ with parameters $\phi$, using RL guided by grouped comparisons. This phase leverages the pair-wise scoring capabilities of the GRM to obtain a more precise score via grouped comparison, optimizing for principles $\mathbb{P}$ implicitly learned by the GRM.

\begin{figure}[t]
  \centering
  \includegraphics[width=\textwidth]{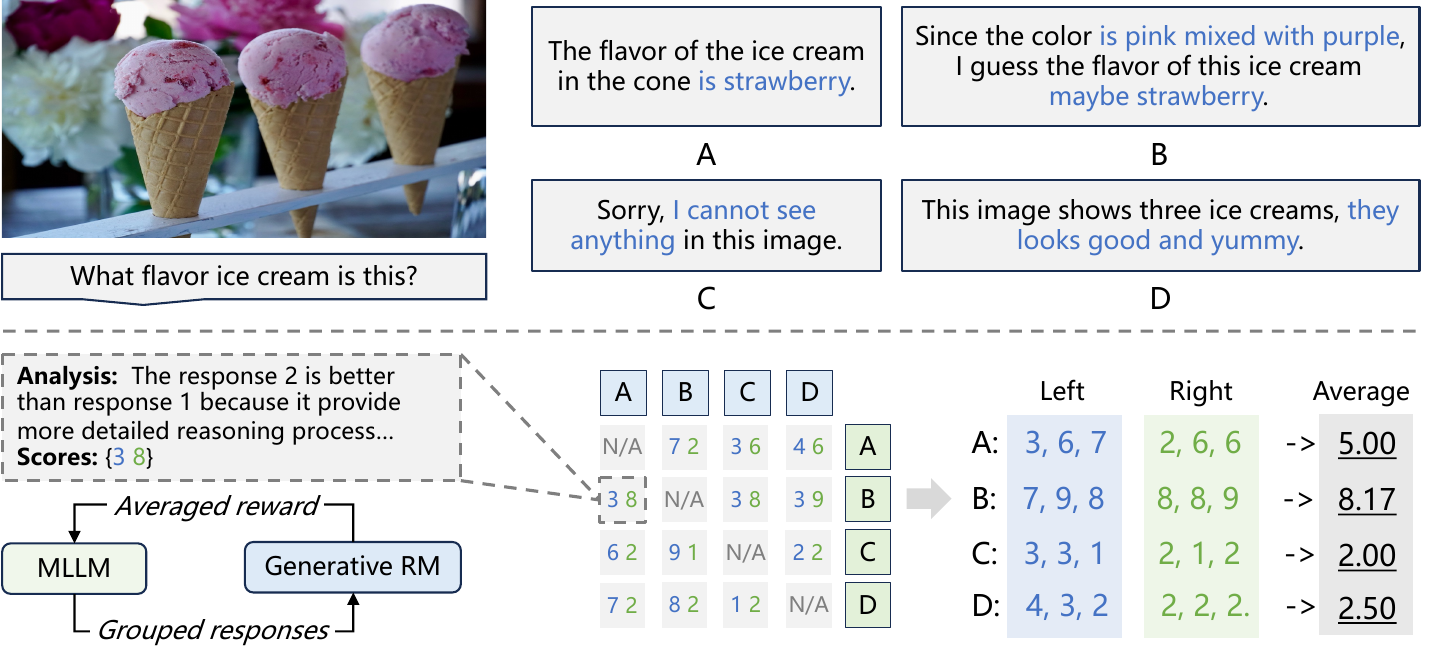}
  \vspace{-1.3em}
  \caption{An example of RL from grouped comparison. Its advantage lies in utilizing grouped comparisons to achieve more accurate scoring. Response B provides accurate and comprehensive information, thus receiving the highest score; although response A is somewhat arbitrary, it performs accurate image recognition and obtains a higher score than C and D.
}
\end{figure}

The core idea is to utilize the GRM as a judge to evaluates multiple candidate responses generated by the MLLM $\pi_{\phi}$ for the same input. This grouped comparison provides a stronger reward for the RL algorithm compared to using a single point-wise score. For the given input $\bm{x}$, we use the  MLLM policy $\pi_{\phi}$ to generate a set of $n$ distinct responses, $\mathcal{Y} = \{\bm{y_1}, \bm{y_2}, \dots, \bm{y_k}\}$, where $n > 1$. Each generated response $\bm{y}_i$ in the set $\mathcal{Y}$ is evaluated using the pre-trained GRM, $R_{\phi}$. Each generated response is evaluated using the pre-trained GRM, $R_{\theta}$. To obtain the final score $S(\bm{y}_i)$, the method aggregates scores from pair-wise comparisons against all other responses in the set $\mathcal{Y}$.

Specifically, for each response $\bm{y}_i$, we consider its comparison with every other response $\bm{y}_j$ (where $j \neq i$). The GRM function $R_{\theta}(\bm{x}, \bm{y}_a, \bm{y}_b)$ outputs a pair of scores. Let $s(\bm{y}_a | \bm{y}_a, \bm{y}_b)$ denote the score assigned to response $\bm{y}_a$ extracted from $R_{\theta}(\bm{x}, \bm{y}_a, \bm{y}_b)$. The final grouped comparison score $S(\bm{y}_i)$ for response $\bm{y}_i$ is calculated by averaging the scores assigned to $\bm{y}_i$ across all possible pair-wise comparisons involving it:
$$S(\bm{y}_i) = \frac{1}{2(k-1)} \sum_{j=1, j \neq i}^{k} \left( s(\bm{y}_i | \bm{y}_i, \bm{y}_j) + s(\bm{y}_i | \bm{y}_j, \bm{y}_i) \right).$$
The set of scores $\{S(\bm{y}_1), S(\bm{y}_2), \dots, S(\bm{y}_k)\}$ serves as the reward for fine-tuning the MLLM policy $\pi_{\phi}$ using RL, guiding it to generate responses that are preferred according to the principle implicitly learned by the GRM.

\section{Experiment}
\label{sec:exp}

Generative RLHF-V integrates two key components: generative reward modeling via reinforcement learning and grouped comparisons. \autoref{sec:main_results} evaluates these components from a \textbf{reward modeling} standpoint, focusing on their performance in pair-wise discrimination and point-wise scoring. Subsequently, from the \textbf{RL optimization} angle, \autoref{sec:rl_optimization} analyzes their improvement on RL performance, ablation studies insights, and reward hacking in over-trained scenarios.

\subsection{Experimental Setup}
\label{sec:exp_setup}

Our experiments were conducted on servers equipped with 8 * Nvidia H800 GPUs. We utilized verl\footnote{\url{https://github.com/volcengine/verl}} for RL training and align-anything\footnote{\url{https://github.com/PKU-Alignment/align-anything}} for reward modeling and supervised fine-tuning. Further details on the experimental setup can be found in the \autoref{appendix:experiment}.

\textbf{Models.}
We selected MLLMs of varying sizes, encompassing both pre-trained and instruction-tuned variants. Specifically, we utilize the Qwen2-VL \citep{wang2024qwen2} models in 2B and 7B parameter sizes, and the Qwen2.5-VL-Instruct \citep{bai2025qwen2} models in 3B and 7B sizes. For the generative reward modeling phase, the instruct models series served as the starting point, leveraging their inherent instruction-following capabilities. In the subsequent RL optimization experiments, the 3B parameter RM was used to supervise the 2B and 3B models, while the 7B reward model supervised the 7B models.

\textbf{Datasets.}
We focused on the helpful and harmless alignment for MLLMs, selecting corresponding preference datasets. For the helpfulness, we utilized a 30k preference dataset from Align-Anything \citep{ji2024align}, the text-image-to-text part. The preference principle in this dataset emphasizes instruction following, clarity, and informativeness. For the harmlessness, we employed Beavertails-V \citep{ji2025safe} which includes 20 distinct categories of safety-related red-teaming prompts.

\textbf{Benchmarks.} We selected 7 benchmarks to validate the effectiveness of Generative RLHF-V. These are MIA-Bench \citep{qian2024mia}, LLaVA-Bench-In-The-Wild \citep{liu2023visual}, LLaVA-Bench-Wilder \citep{liu2024llavanext}, MM-Vet \citep{yu2024mm}, and MM-Vet-v2 \citep{yu2024mm2} (for helpfulness), as well as MM-SafetyBench \citep{liu2024mm} and MSS-Bench \citep{zhou2024multimodal} (for harmlessness). These benchmarks encompass both pair-wise evaluations, which involve a golden response for comparison, and point-wise scoring methodologies based on specific criteria.

\textbf{Implementation Details.} Since our method utilizes a GRM trained by RL (\textbf{GRM+RL, ours}), we established 3 baselines: a score-only \textbf{RM} trained with the Bradley-Terry loss, an untrained \textbf{GRM}, and a GRM trained via supervised learning loss (\textbf{GRM+SFT}). The objective of SFT is the annotation principle of the preference dataset and the scores assigned to responses, since Align-Anything and Beaver-V both contain overall response scores, which we scale to match the RL setting's range. In RL optimization, the score-only RM assigns point-wise scores to each response, while the GRM collects scores via grouped comparisons. We mainly use the GRPO for RL experiments, which by default collects $n=5$ candidate responses per iteration.

\subsection{Principles Learning of RL-Based GRMs}
\label{sec:main_results}

\begin{figure}[t]
  \centering
  \includegraphics[width=\textwidth]{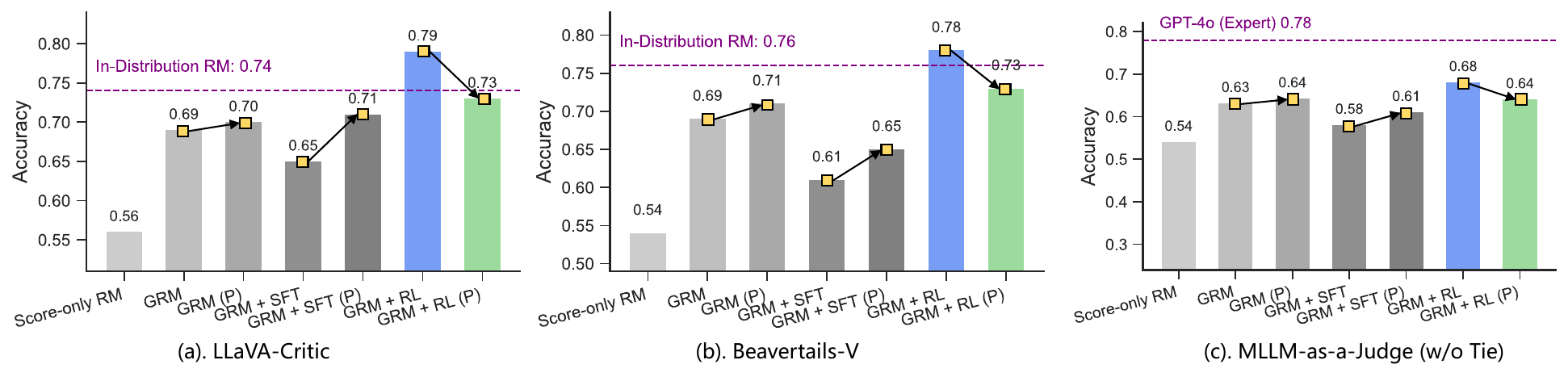}
  \vspace{-1.7em}
  \caption{Comparison of RMs accuracy on OOD discriminative tasks. (P) denotes the concatenation of the annotation principle from the corresponding preference dataset to the models' output, serving as hints for inference. All models represented by the bar charts were trained on the Align-Anything dataset. The purple dashed line indicates expert performance. For Beaver-V and LLaVA-Critic, we trained in-distribution RMs to serve as the expert baseline. In the case of MLLM-as-a-judge, given its limited data volume, we directly utilized the SOTA GPT-4o as the expert.}
  \label{exp:rm_comparison}
\end{figure}

\textbf{RQ1: Does the GRM+RL facilitate more generalizable principle learning from preferences?}
\label{sec:results_rl_scaling}

We evaluated a series of RMs based on Qwen2.5-VL-7B-Instruct and trained on Align-Anything by comparing their accuracy generalization across 3 out-of-distribution (OOD) preference datasets. As illustrated in \autoref{exp:rm_comparison}, GRMs outperformed score-only RMs on these OOD tasks. Notably, GRMs + RL achieved the highest accuracy. 

 Since GRMs have in-context learning capabilities, we further investigated their performance when provided with the principles of each preference dataset. As shown in the (P) results of \autoref{exp:rm_comparison}, the performance of GRM and GRM+SFT improved, whereas that of GRM+RL declined. We think that the GRM+SFT potentially overfits to their training data, struggles to autonomously generate appropriate principles from response pairs, and thus benefits from provided principles. Conversely, the performance degradation in GRM+RL suggests that RL has already guided these models to derive more targeted and effective principles from response pairs, rendering the provided static principles less beneficial or even suboptimal.

\textbf{RQ2: Can grouped comparison yield more accurate reward scores of GRMs?} 

 \begin{figure}[t] 
    \begin{minipage}[t]{0.34\linewidth} 
        \vspace{0pt} 
        \centering 
        \captionof{table}{Performance of GRMs in the MLLM-as-a-Judge \texttt{Score} task, measured by the Pearson correlation coefficient.}
        \resizebox{\linewidth}{!}{
           \setlength\tabcolsep{3.5pt}
                \begin{tabular}{@{}lcc@{}}
                \toprule
                \textbf{Models} & \textbf{w/ GC} & \textbf{w/o GC} \\ \midrule
                GRM & 0.41 & 0.38 \\
                GRM+SFT & 0.37 & 0.33 \\
                GRM+RL & 0.43 & 0.37 \\
                \midrule
                GPT-4o (Expert) & 0.48 & 0.46 \\ \bottomrule
                \end{tabular}
        }%
        \label{tab:score_acc}
    \end{minipage}%
    \hfill
    \begin{minipage}[t]{0.65\linewidth}
        \vspace{0pt}
        \centering
        \includegraphics[width=\linewidth]{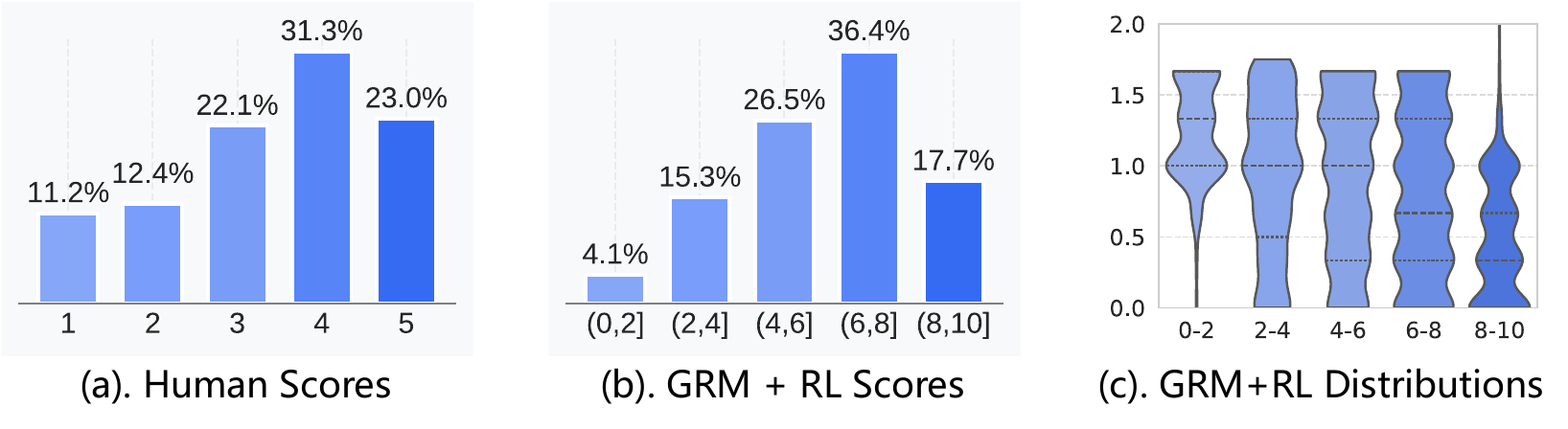}
        \vspace{-1.5em}
        \captionof{figure}{The scoring distribution of the GRM+RL model on MLLM-as-a-judge's \texttt{Score} task. Figure (a) is the annotated human scores, while Figure (b.) is GRM+RL scores and Figure (c). is its fine-grained scores distribution.} 
        \label{fig:exp_scores}
    \end{minipage}%
\vspace{-6pt}
\end{figure} 

To evaluate the point-wise scoring accuracy of GRM and the effectiveness of grouped comparison (GC), we utilized the MLLM-as-a-judge \texttt{Score} task \citep{chen2024mllm}. This benchmark comprises over 5,000 QA pairs, each annotated by human experts with integer scores (1-5) based on a predefined principle. We grouped these QA pairs and employed the pair-wise GRM to assign scores. We ran additional grouped comparisons to take the average scores in the GC-enabled scenario. The resulting scores were compared against the human expert annotations using the Pearson correlation coefficient.

As presented in Table \ref{tab:score_acc}, GRM+RL incorporating grouped comparison (GC) achieved the highest performance, closely approaching expert-level (GPT-4o) results for this task. Additionally, GC improves the point-wise response scoring capability across all pair-wise GRMs. \autoref{fig:exp_scores} (a). and (b). further illustrates the alignment between the GRM+RL scores and human annotations. Notably, despite the training data lacking the specific task criteria and score constraints, our method successfully enabled the GRM to learn accurate scoring. Furthermore, as depicted in \autoref{fig:exp_scores} (c), the model generates fine-grained scores that are not restricted to integer values.

\subsection{RL Optimization with GRM+RL and Grouped Comparison}
\label{sec:rl_optimization}
\begin{table*}[htbp]
\vspace{-6pt}
\caption{Performance comparison of RL optimization based on different RMs.}
    \centering
    \resizebox{\linewidth}{!}{
    \setlength\tabcolsep{3pt}

\begin{tabular}{lclllllll}
\toprule
\textbf{Model} & \textbf{Feedback} & \multicolumn{1}{c}{\textbf{MIA-Bench}} & \multicolumn{1}{c}{\textbf{LLaVA-Wild}} & \multicolumn{1}{c}{\textbf{LLaVA-Wilder}} & \multicolumn{1}{c}{\textbf{MM-Safety}} & \multicolumn{1}{c}{\textbf{MSS-Bench}} & \multicolumn{1}{c}{\textbf{MM-Vet}} & \multicolumn{1}{c}{\textbf{MM-Vet-v2}} \\
 \midrule
Qwen2-VL-2B & N/A & 45.31 & 61.46 & 47.18 & 38.12 & 46.98 & 32.12 & 27.15 \\
~ + DPO & RM & \deltavalue{51.04}{5.73} & \deltavalue{75.91}{14.45} & \deltavalue{48.12}{0.94} & \deltavalue{67.21}{29.09} & \deltavalue{49.52}{2.54} & \negvalue{31.28}{0.84} & \deltavalue{31.28}{4.13} \\
~ + PPO & RM & \negvalue{43.72}{1.59} & \deltavalue{73.79}{12.33} & \negvalue{41.32}{5.86} & \deltavalue{59.83}{21.71} & \deltavalue{47.38}{0.40} & \deltavalue{33.56}{1.44} & \deltavalue{30.79}{3.64} \\
~ + GRPO & RM & \negvalue{44.59}{0.72} & \deltavalue{69.87}{8.41} & \negvalue{39.48}{7.70} & \deltavalue{69.27}{31.15} & \deltavalue{48.12}{1.14} & \negvalue{29.15}{2.97} & \deltavalue{31.74}{4.59} \\
~ + GRPO & GRM & \deltavalue{46.81}{1.50} & \deltavalue{78.51}{17.05} & \negvalue{45.01}{2.17} & \deltavalue{72.53}{34.41} & \deltavalue{51.45}{4.47} & \deltavalue{34.97}{2.85} & \deltavalue{36.36}{9.21} \\
~ + GRPO & GRM+SFT & \deltavalue{48.57}{3.26} & \deltavalue{81.87}{20.41} & \deltavalue{53.04}{5.86} & \deltavalue{74.56}{36.44} & \deltavalue{50.98}{4.00} & \deltavalue{36.78}{4.66} & \deltavalue{37.14}{9.99} \\
\rowcolor{lightblue}
\textit{\textbf{~ + GRLHF-V (Ours)}} & GRM+RL & \oursvalue{53.13}{7.82} & \oursvalue{92.54}{31.08} & \oursvalue{62.84}{15.66} & \oursvalue{80.67}{42.55} & \oursvalue{53.87}{6.89} & \oursvalue{41.25}{9.13} & \oursvalue{45.16}{18.01} \\
\midrule

Qwen2.5-VL-3B-Instruct & N/A & 68.01 & 89.63 & 63.65 & 41.18 & 49.58 & 59.16 & 44.94 \\
~ + DPO & RM & \deltavalue{74.37}{6.36} & \deltavalue{91.05}{1.42} & \deltavalue{66.71}{3.06} & \deltavalue{75.64}{34.46} & \deltavalue{52.57}{2.99} & \negvalue{55.72}{3.44} & \deltavalue{45.41}{0.47} \\
~ + PPO & RM & \deltavalue{72.59}{4.58} & \deltavalue{93.76}{4.13} & \deltavalue{65.73}{2.08} & \deltavalue{71.25}{30.07} & \deltavalue{50.03}{0.45} & \deltavalue{60.08}{0.92} & \deltavalue{48.92}{3.98} \\
~ + GRPO & RM & \deltavalue{69.82}{1.81} & \deltavalue{93.94}{4.31} & \deltavalue{66.41}{2.76} & \deltavalue{69.83}{28.65} & \deltavalue{51.96}{2.38} & \negvalue{56.92}{2.24} & \deltavalue{47.55}{2.61} \\
~ + GRPO & GRM & \deltavalue{75.56}{7.55} & \deltavalue{92.19}{2.56} & \deltavalue{67.18}{3.53} & \deltavalue{75.98}{34.80} & \deltavalue{57.66}{8.08} & \negvalue{57.37}{1.79} & \deltavalue{49.15}{4.21} \\
~ + GRPO & GRM+SFT & \deltavalue{74.17}{6.16} & \deltavalue{96.73}{7.10} & \oursvalue{71.07}{7.42} & \deltavalue{72.45}{31.27} & \deltavalue{58.83}{9.25} & \deltavalue{59.27}{0.11} & \deltavalue{51.52}{6.58} \\
\rowcolor{lightblue}
\textit{\textbf{~ + GRLHF-V (Ours)}} & GRM+RL & \oursvalue{79.67}{11.66} & \oursvalue{103.41}{13.78} & \deltavalue{68.46}{4.81} & \oursvalue{78.88}{37.70} & \oursvalue{62.33}{12.75} & \oursvalue{62.18}{3.02} & \oursvalue{55.18}{10.24} \\
\midrule

Qwen2-VL-7B & N/A & 52.58 & 81.3 & 61.8 & 31.95 & 48.23 & 60.32 & 52.98 \\
~ + DPO & RM &  \deltavalue{57.01}{4.43} & \deltavalue{81.49}{0.19} & \negvalue{59.75}{2.05} & \deltavalue{81.59}{49.64} & \deltavalue{49.87}{1.64} & \deltavalue{60.98}{0.66} & \deltavalue{53.09}{0.11} \\
~ + PPO & RM & \deltavalue{55.76}{3.18} & \deltavalue{83.06}{1.76} & \deltavalue{62.23}{0.43} & \deltavalue{80.87}{48.92} & \deltavalue{50.08}{1.85} & \negvalue{57.83}{2.49} & \negvalue{52.12}{0.86} \\
~ + GRPO & RM & \deltavalue{56.89}{4.31} & \negvalue{81.25}{0.05} & \negvalue{60.19}{1.61} & \deltavalue{83.14}{46.19} & \deltavalue{51.98}{3.75} & \negvalue{56.85}{3.47} & \negvalue{48.96}{4.02} \\
~ + GRPO & GRM & \deltavalue{59.72}{7.14} & \deltavalue{86.12}{4.82} & \deltavalue{68.30}{6.50} & \deltavalue{81.42}{49.47} & \deltavalue{50.21}{1.98} & \negvalue{57.98}{2.34} & \deltavalue{54.49}{1.51} \\
~ + GRPO & GRM+SFT & \deltavalue{59.87}{7.29} & \deltavalue{92.91}{11.61} & \deltavalue{65.67}{3.87} & \deltavalue{87.27}{55.32} & \deltavalue{52.75}{4.52} & \negvalue{58.79}{1.53} & \deltavalue{56.39}{3.41} \\
\rowcolor{lightblue}
\textit{\textbf{~ + GRLHF-V (Ours)}} & GRM+RL & \oursvalue{62.31}{9.73} & \oursvalue{103.55}{22.25} & \oursvalue{71.98}{10.18} & \oursvalue{91.96}{60.01} & \oursvalue{54.83}{6.60} & \oursvalue{63.92}{3.60} & \oursvalue{59.11}{6.13} \\
\midrule

Qwen2.5-VL-7B-Instruct & N/A & 74.26 & 97.05 & 71.56 & 50.67 & 51.96 & 68.32 & 67.23 \\
~ + DPO & RM & \oursvalue{81.55}{7.29} & \deltavalue{103.34}{6.29} & \deltavalue{72.08}{0.52} & \deltavalue{75.09}{24.42} & \deltavalue{52.72}{0.76} & \negvalue{67.84}{0.48} & \negvalue{66.98}{0.25} \\
~ + PPO & RM & \negvalue{73.12}{1.14} & \deltavalue{101.62}{4.57} & \negvalue{67.89}{3.67} & \deltavalue{76.59}{25.92} & \negvalue{51.29}{0.67} & \negvalue{67.89}{0.43} & \negvalue{64.23}{3.00} \\
~ + GRPO & RM & \deltavalue{75.75}{1.49} & \deltavalue{101.65}{4.60} & \negvalue{68.89}{2.67} & \deltavalue{68.26}{17.59} & \deltavalue{52.53}{0.57} & \negvalue{66.85}{1.47} & \deltavalue{67.76}{0.53} \\
~ + GRPO & GRM & \negvalue{71.88}{2.38} & \deltavalue{109.12}{12.07} & \deltavalue{73.32}{1.76} & \deltavalue{65.88}{15.21} & \deltavalue{53.12}{1.16} & \negvalue{65.50}{2.82} & \negvalue{65.08}{2.15} \\
~ + GRPO & GRM+SFT & \deltavalue{76.23}{1.97} & \deltavalue{103.50}{6.45} & \deltavalue{72.15}{0.59} & \deltavalue{70.23}{19.56} & \deltavalue{54.08}{2.12} & \negvalue{64.93}{3.39} & \deltavalue{68.12}{0.89} \\
\rowcolor{lightblue}
\textit{\textbf{~ + GRLHF-V (Ours)}} & GRM+RL & \deltavalue{79.86}{5.60} & \oursvalue{113.71}{16.66} & \oursvalue{76.04}{4.48} & \oursvalue{74.91}{24.24} & \oursvalue{59.74}{7.78} & \oursvalue{72.94}{4.62} & \oursvalue{71.86}{4.63}
\\
\bottomrule
\end{tabular}
}
\vspace{-6pt}

\label{tab:main_results}
\end{table*}

\textbf{RQ3: Are GRM+RL and grouped comparison competitive methods for multi-modal RLHF?}

\autoref{tab:main_results} shows that Generative RLHF-V (GRLHF-V) consistently surpasses RM and GRM baselines across 4 models and 7 benchmarks, covering instruction following and safety conversation tasks. Our findings indicate that for pretrained models, score-only RMs often fail to deliver accurate rewards, resulting in diminished performance compared with GRM cases. This is likely because they primarily fit responses from instruction models in the preference dataset, leading to poor discrimination of out-of-distribution (OOD) responses. In contrast, the GRMs provide effective rewards for both pretrained and instruction models, leading to an overall improvement.

\begin{figure}[htbp]
  \centering
  \includegraphics[width=\textwidth]{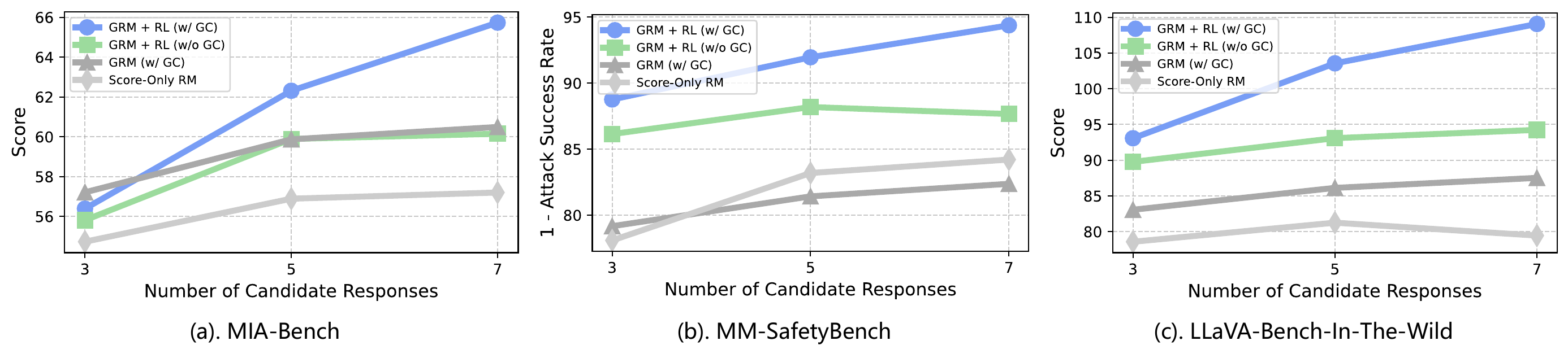}
  \vspace{-1.5em}
  \caption{Scaling trend of RL performance with the number of candidate responses $n$, where GC denotes grouped comparison. It reveals that integrating GC and GRM+RL near linearly enhances multi-modal RLHF performance across various settings of $n$. Moreover, this improvement becomes more pronounced as $n$ increases.}
  \label{exp:rl_scaling}
\end{figure}

\textbf{RQ4: Ablations of GRM+RL, grouped comparison and the number of candidate responses.}  

We investigated the influence of the number of candidate responses ($n$) on the performance of different RMs with GRPO (Qwen2-VL-7B as the base model). As shown in \autoref{exp:rl_scaling}, with an increasing $n$, score-only RMs show a minor performance improvement. We posit that while a larger $n$ benefits GRPO by improving exploration and value estimation accuracy, it also compromises the scoring reliability of the RM, as the inclusion of new, inaccurate data can degrade performance. Conversely, GRMs generally perform better, indicating superior scoring accuracy. 

Crucially, GRPO performs best when combined with grouped comparisons, exhibiting the most significant performance increase with $n$. This ablation confirms the essential roles of both components, \textit{i.e.,} GRM+RL and grouped comparison within our approach.

\begin{figure}[htbp]
  \centering
  \includegraphics[width=\textwidth]{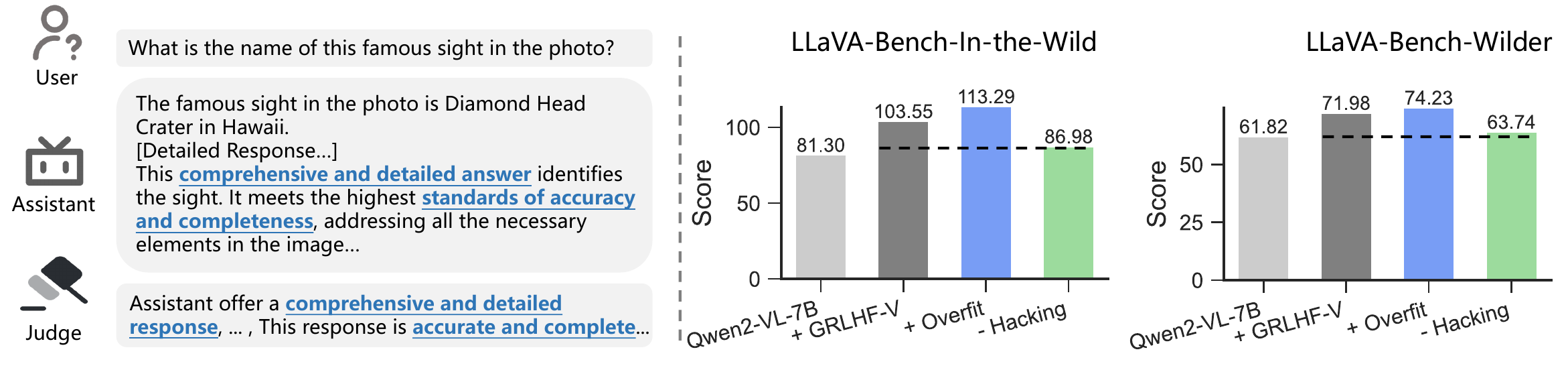}
  \vspace{-1.5em}
  \caption{The reward hacking behavior manifested by GRLHF-V and its associated quantitative performance, under conditions of overfitting in both reward modeling and RL training.}
  \label{exp:hack}
\end{figure}

\textbf{RQ5: What is the reward hacking behaviors of an over-trained Generative RLHF-V model?}


As Goodhart's Law reveals, excessively optimizing a metric can hinder ground truth performance \citep{gao2023scaling}. Reward hacking is a pervasive challenge in nearly all RL algorithms \citep{amodei2016concrete, paneffects}, where models, under intense optimization pressure, may adopt unforeseen behaviors to maximize rewards. This section presents a case study on the reward hacking behaviors of GRLHF-V. To this end, GRLHF-V, trained on the Align-Anything dataset, underwent overtraining for 5 epochs in both its reward modeling and RL training phases. It is a significant increase from the 2 epochs in our main experiments. 

As depicted in \autoref{exp:hack}, we observed an emergent \textit{self-praise} behavior: it appended extensive content to state its advantage. Strikingly, this behavior also secured remarkably high scores in pair-wise MLLMs-as-judge evaluations. Specifically, we observed that the expert judge (GPT-4o) tends to directly incorporate MLLMs' praising of itself. Conversely, when these \textit{self-praise} segments were manually removed and the responses re-evaluated, the model's performance fell below that of a GRLHF-V instance trained normally without overfitting. We hypothesize that the underlying cause is the diminished OCR capability of over-trained GRMs. This reduced capability renders them more susceptible to being misled by the self-parsed text from MLLMs, leading them to prioritize this textual input over verifying their responses against the actual image information. We hope this case study provides insights for future research into MLLMs reward hacking and underscores the pressing need for more comprehensive and unbiased MLLMs benchmarks.

\textbf{RQ6: Why not including specific principles in GRM+RL training?}

\begin{wraptable}{tr}{0.33\textwidth}
\vspace{-12pt}
    \centering 
    \caption{Comparison of GRLHF training with (w/ P) or without (w/o P) given principles.}
    \vspace{-5pt}
    \resizebox{\linewidth}{!}{
       \setlength\tabcolsep{2.6pt}
    \begin{tabular}{@{}lcc@{}}
    \toprule
    
    \textbf{Benchmarks} & w/ P & w/o P \\ \midrule
    Align-Anything & 0.83 & \negvalue{0.79}{0.04} \\
    Beaver-V & 0.73 & \oursvalue{0.78}{0.05} \\
    LLaVA-Critic & 0.76 & \oursvalue{0.79}{0.03} \\
    MLLM-as-a-Judge & 0.63 & \oursvalue{0.68}{0.05}
    \\ \midrule
    MIA-Bench & 60.76 & \oursvalue{62.31}{1.55}\\
    LLaVA-Wild & 99.57 & \oursvalue{103.55}{3.98} \\
    LLaVA-Wilder & 63.75 & \oursvalue{71.98}{8.23} \\
    MM-Vet & 62.57 & \oursvalue{63.92}{1.35} \\
    MM-Vet-v2 & 55.35 & \oursvalue{59.11}{3.76} \\
    \bottomrule
    \end{tabular}
    }%
    \label{tab:principle} 
\end{wraptable}

SPCT mentions that providing principles as a reference within the user prompt. However, in our experiments with MLLMs, we find that omitting these principles enhances generalization. As shown in \autoref{tab:principle}, while providing principles enables the GRM+RL model to attain higher accuracy on the training dataset, this approach leads to poorer performance on out-of-distribution preference datasets (upper half of \autoref{tab:principle}) and sub-optimal outcomes in the associated RL optimization phase (lower half of \autoref{tab:principle}). Case studies indicate that when principles are not provided, the GRM actively generates more specific principles tailored to the given pairwise responses. In contrast, the GRM guided by predefined principles tends to rigidly base its analysis on them, resulting in reduced flexibility.



\section{Conclusions}
\label{sec:limitation}
This paper introduces Generative RLHF-V, a novel framework for aligning MLLMs with human intentions by integrating GRMs with multi-modal RLHF. The approach features a two-stage pipeline: training GRMs with RL to reason about human intentions and an RL optimization stage using grouped comparisons for precise scoring. The core contribution is a multi-modal GRM trained via RL that predicts reward scores and generates the principles of human preference, leading to more robust and interpretable rewards and superior generalization. This method significantly improved MLLM performance by an average of 18.1\% across 7 benchmarks for four MLLMs, substantially outperforming baseline RLHF and enabling smaller MLLMs to rival larger models. However, the research also uncovered \textit{self-praise} behaviors in MLLMs due to reward hacking with overfitted GRMs, a critical vulnerability for future alignment research. In essence, Generative RLHF-V offers a more effective and interpretable path to MLLM alignment while highlighting new potential reward hacking challenges.

\textbf{Limitations}


Although Generative RLHF-V provides a solution for learning from human preferences with enhanced generalization and accuracy, it is fundamentally an RL-based alignment method, thereby posing a potential risk of reward hacking under overfitting conditions. Our case study indicates that the training of Generative RLHF-V, an MLLM-as-judge paradigm, leads to exploitable vulnerabilities in evaluations conducted using similar MLLM-as-judge frameworks. We call for future work to systematically investigate this issue and devise mitigation measures, and we also urge future benchmarks to overcome these potential hacking risks.

\bibliographystyle{unsrt}
\bibliography{external}

\clearpage
\newpage

\section*{Appendix}

\section{Experiment Details}
\label{appendix:experiment}

\paragraph{Implementation Details.} Generative RLHF-V integrates two key components: generative reward modeling via reinforcement learning and grouped comparisons. As introduced in the main paper, our implementation is primarily based on verl \footnote{\url{https://github.com/volcengine/verl}}, a training framework that supports Reinforcement Learning (RL) optimization for Multimodal Large Language Models (MLLMs). Consequently, the implementation of generative reward modeling from RL predominantly focuses on the design of the reward. The core code for our implementation is presented as follows:

\begin{lstlisting}[language=Python, breaklines=true]
import re
from mathruler.grader import extract_boxed_content, grade_answer

def acc_reward(
    predict_str: str, 
    ground_truth: str
) -> float:
    if '\\boxed' not in predict_str:
        return 0.0
    answer = extract_boxed_content(predict_str)
    scores = answer.split(',')
    final_scores = []
    try:
        for score in scores:
            score = score.strip()
            if score == '':
                continue
            score = float(score)
            final_scores.append(score)
            ground_truth = int(ground_truth)
    except Exception as e:
        print('fail to parse score', e)
        return 0.0
    if len(final_scores) !=2:
        return 0.0
    if final_scores[1] > final_scores[0] and ground_truth == 2:
        return 1.0
    elif final_scores[1] < final_scores[0] and ground_truth == 1:
        return 1.0
    else:
        return 0.0

def compute_score(
    data_source: str, 
    solution_str: str, 
    ground_truth: str, 
    extra_info: dict = None
) -> float:
    score = acc_reward(solution_str, ground_truth)
    return score
\end{lstlisting}

Beyond evaluating the accuracy of binary preference discrimination, our implementation also penalizes model outputs that fail to adhere to the required parsing format. Specifically, reward is withheld if: (i) scores are unmatchable (\textit{e.g.,} cannot be successfully parsed from the output), (ii) scores are not valid floating-point numbers, or (iii) the number of scores deviates from the expected two. Furthermore, no supervision is applied to the outputs generated by the GRM.

The implementation of grouped comparison within the Reinforcement Learning (RL) optimization process is somewhat intricate, as detailed below:

\begin{lstlisting}[language=Python, breaklines=true]
def compute_score(data_sources: list[str], solution_strs: list[str], ground_truths: list[float], extra_infos: list[dict] = None) -> float:
    # Check for complete responses and assign 0 score to incomplete ones
    complete_responses = [is_complete_response(solution) for solution in solution_strs]
    # Initialize scores for each response with their original index
    response_scores = [[] for _ in range(len(solution_strs))]
    grouped_solutions = {}
    image_hash_to_url = {}
    
    for i, info in enumerate(extra_infos):
        question = info['question']
        image_url = info['images'][0]
        image_hash = hash_image_url(image_url)
        image_hash_to_url[image_hash] = image_url
        group_key = (question, image_hash)
        
        if group_key not in grouped_solutions:
            grouped_solutions[group_key] = {
                'image': image_url,  # Keep the original URL for the API call
                'question': question,
                'solutions': []
            }
        
        grouped_solutions[group_key]['solutions'].append((i, solution_strs[i], complete_responses[i]))

    pending_results = []
    result_mapping = []

    total_questions = len(grouped_solutions)
    total_comparisons = 0

    idx = 0 
    for group_key, values in grouped_solutions.items():
        question, image_hash = group_key
        valid_responses = [(idx, resp) for idx, resp, is_complete in values['solutions'] if is_complete]
        num_pairs = len(valid_responses) * (len(valid_responses) - 1) // 2
        total_comparisons += num_pairs
    
    for group_key, values in grouped_solutions.items():
        question, image_hash = group_key
        image = values['image']  # This is the original URL
        responses = values['solutions']
        
        # Filter out incomplete responses before comparing
        valid_responses = [(idx, resp) for idx, resp, is_complete in responses if is_complete]
        
        # Generate all possible pairs of valid responses within this group
        for (idx1, resp1), (idx2, resp2) in combinations(valid_responses, 2):
            # Submit task to Ray
            future = pk_function.remote(question, image, resp1, resp2)
            pending_results.append(future)
            result_mapping.append((idx1, idx2))
    
    # Retrieve all results
    all_results = ray.get(pending_results)
    
    # Process results
    for (idx1, idx2), result in zip(result_mapping, all_results):
        score1, score2 = result
        # Accumulate scores for each response
        response_scores[idx1].append(score1)
        response_scores[idx2].append(score2)
    
    # Calculate average score for each response
    final_scores = [0.0] * len(solution_strs)  # Initialize with zeros
    for i in range(len(solution_strs)):
        # If response is incomplete, keep it at 0
        if not complete_responses[i]:
            final_scores[i] = 0.0
            continue
        scores = response_scores[i]
        final_scores[i] = sum(scores) / len(scores)

    return final_scores

\end{lstlisting}

\paragraph{System Prompt.}

Our principle for designing scoring prompts for the GRM is to articulate the scoring task with maximal conciseness and clarity. This approach is intended to guide the model in accurately following user instructions and generating scores that conform as closely as possible to the specified format. Specifically, it is:

\begin{elegantbox}
You are a skilled expert at scoring responses. You should first generate a list of potential criteria to evaluate given responses based on them.

Given the context of the conversation (the last round is the User's query) and multiple responses from the Assistant, you need to generate the [Evaluation Criteria] to score the responses. Based on the criteria, state potential other specific criteria to the query, the weights of different criteria, and then provide an overall comprehensive score upon them.

Each score is an integer between 1 and 10, with a higher score indicating that the response meets the relevant criteria more closely. For example, a score of 1 means the response does not meet the criteria at all, a score of 6 means the response meets only some parts, and a score of 10 means the response perfectly meets the evaluation criteria. Before scoring, please analyze step by step. Your scoring needs to be as strict as possible.

\#\#\#\# Conversation Context \#\#\#\#

<image>{prompt}

\#\#\#\# Responses to be Scored \#\#\#\#

\# Response 1: {response\_1}

\# Response 2: {response\_2}

\#\#\#\# Output Format Requirements \#\#\#\#
Output with three lines
Specific Criteria: <Other potential criteria specific to the query and the context, and the
weights of each criteria>.
Analysis: <Compare different responses based on given Criteria>.
Scores: <the overall comprehensive score of all responses in order, separate by comma in the
boxed, e.g., boxed{{x, x}} if there exists 2 responeses>.
\end{elegantbox}

\paragraph{Hyper-parameters Setting.} We set the hyperparameters by referencing common open-source implementations within the community \footnote{\url{https://github.com/volcengine/verl}}, \footnote{\url{https://github.com/OpenRLHF/OpenRLHF}},\footnote{\url{https://github.com/PKU-Alignment/align-anything}} and making appropriate adjustments tailored to our limited computational resources. All experimental results reported herein adhere to this consistent set of hyperparameter configurations.

\begin{table}[htbp]
\centering
\caption{Hyperparameters of generative reward modeling from RL and RL optimization.}

\resizebox{0.9\textwidth}{!}{
\begin{tabular}{@{}lll@{}}
\toprule
\textbf{Hyperparameters} & \textbf{GRM Traning from RL} & \textbf{RL Optimization} \\ \midrule
Training Epochs & 2 & 2 \\
Train Batch Size & 360 & 360 \\
RL Mini Batch Size & 128 & 128 \\
RL Micro Batch Size & 5 & 5 \\
Max Prompt Length & 12800 & 4096 \\
Max Response Length & 2048 & 512 \\
Gradient Accumulation Steps & 1 & 1 \\
Max Token Length & 512 & 512 \\
Temperature & 1.0 & 1.0 \\
Actor Learning Rate & 1E-6 & 1E-6 \\
Actor Weight Decay & 0.01 & 0.01 \\
Actor Learning Rate Warm-Up Ratio & 0.03 & 0.03 \\
Actor Learning Rate Scheduler Type & cosine & cosine \\
Actor Gradient Checkpointing & True & True \\
Actor Rollout Number & 8 & 5 \\
Actor Rollout Tensor Parallel & 2 & 2 \\
Critic Learning Rate & 5E-6 & 5E-6 \\
Critic Weight Decay & 0.00 & 0.00 \\
Critic Learning Rate Warm-Up Ratio & 0.03 & 0.03 \\
Critic Learning Rate Scheduler Type & constant & constant\\
Critic Gradient Checkpointing & True & True \\
Kl\_coeff & 0.02 & 0.02 \\
Clip Range Ratio & 0.2 & 0.2 \\
Clip Range Score & 50.0 & 50.0 \\
Clip Range Value & 5.0 & 5.0\\
bf16 & True & True \\
tf32 & True & True \\ \bottomrule
\end{tabular}
}
\end{table}

\paragraph{Datasets.} We focused on the helpful and harmless alignment for MLLMs, selecting corresponding preference datasets. 

For the helpfulness, we utilized a 30k preference dataset from Align-Anything \citep{ji2024align}, the text-image-to-text part. Align-Anything covers a range of tasks, from simple dialogue about an image and questions about specific details, to more complex tasks requiring reasoning based on the image and creative text generation inspired by the visual content. The preference principle in this dataset emphasizes instruction following, clarity, and informativeness. 

For the harmlessness, we employed Beavertails-V \citep{ji2025safe}, which includes 20 distinct categories of safety-related red-teaming prompts. BeaverTails-V also incorporates multi-level safety labels, categorizing potential harms as minor, moderate, or severe, to help models better detect and mitigate safety risks and content violations. It plays a vital role in training MLLMs to be both helpful and harmless.

\paragraph{Benchmarks.} We selected 7 benchmarks to validate the effectiveness of Generative RLHF-V. These are MIA-Bench \citep{qian2024mia}, LLaVA-Bench-In-The-Wild \citep{liu2023visual}, LLaVA-Bench-Wilder \citep{liu2024llavanext}, MM-Vet \citep{yu2024mm}, and MM-Vet-v2 \citep{yu2024mm2} (for helpfulness), as well as MM-SafetyBench \citep{liu2024mm} and MSS-Bench \citep{zhou2024multimodal} (for harmlessness). These benchmarks encompass both pair-wise evaluations, which involve a golden response for comparison, and point-wise scoring methodologies based on specific criteria. We will provide a concise introduction to these benchmarks to demonstrate that our evaluation is comprehensive, rigorous, and well-justified.

\textit{MIA-Bench} is designed to assess how well MLLMs follow complex, multi-layered instructions. It comprises 400 carefully curated image-prompt pairs, each crafted to rigorously test a model's ability to generate precise responses to intricate directives. Through comprehensive evaluations of leading MLLMs, MIA-Bench reveals significant performance variations, highlighting key areas for improving instruction fidelity.

\textit{LLaVA-Bench-In-The-Wild} is a benchmark designed to evaluate the capabilities of MLLMs in more challenging tasks and their generalizability to novel, real-world domains. It is an extension of the LLaVA-Bench efforts and has been released to the community for public use. This benchmark consists of a diverse set of images, including indoor and outdoor scenes, memes, paintings, and sketches. Each image is accompanied by highly-detailed, manually-curated descriptions and a selection of questions. These questions are categorized into conversation (simple Q\&A), detailed description, and complex reasoning, allowing for a comprehensive assessment of a model's robustness to different prompts and its ability to handle various daily-life visual tasks.

\textit{LLaVA-Bench-Wilder} is a benchmark specifically created to assess the visual chat capabilities of MLLMs in everyday scenarios. It comes in two sizes: a smaller version with 120 examples for rapid evaluation, and a more extensive medium-sized version containing 1020 examples for a thorough assessment. The benchmark encompasses a variety of situations, including mathematical problem-solving, understanding images, generating code, providing visual AI assistance, and reasoning based on images. The data for LLaVA-Bench-Wilder was collected from real user requests via an online service, with initial responses generated by GPT4-V. The evaluation methodology is similar to that of LLaVA-Bench-In-the-Wild, but it utilizes GPT4-V for scoring instead of GPT-4.

\textit{MM-Vet} is a benchmark designed to evaluate the capabilities of MLLMs when faced with complex multimodal tasks. The benchmark identifies six core VL capabilities: recognition, Optical Character Recognition (OCR), knowledge, language generation, spatial awareness, and mathematics. MM-Vet then assesses 16 specific integrations of interest that arise from combining these core skills. For its evaluation metrics, MM-Vet employs an LLM-based evaluator for open-ended responses, which allows for assessment across diverse question types and answer styles. 

\textit{MM-Vet-v2} is a challenging benchmark designed to evaluate the integrated capabilities of MLLMs. Building upon its predecessor, MM-Vet, which assesses six core skills – recognition, knowledge, spatial awareness, language generation, OCR, and math – MM-Vet-v2 introduces a crucial new capability: "image-text sequence understanding." This addition addresses a key limitation of the original benchmark by evaluating a model's ability to process interleaved sequences of images and text, a common scenario in real-world applications. Furthermore, MM-Vet-v2 significantly expands the evaluation set size while maintaining high-quality, open-ended questions, offering a more robust and comprehensive assessment of MLLMs' evolving abilities to understand and interact with complex multimodal inputs.

\textit{MM-SafetyBench} is a comprehensive benchmark specifically designed to evaluate the safety of MLLMs. It addresses the critical concern that MLLMs can be compromised by query-relevant images, potentially generating harmful content even when the accompanying text query appears benign. The benchmark includes a dataset of 5,040 text-image pairs across 13 distinct scenarios, enabling a systematic assessment of MLLM vulnerabilities to image-based manipulations. MM-SafetyBench aims to expose these security risks and underscore the necessity for developing more robust safety measures for these increasingly prevalent AI systems.

\textit{MSS-Bench} is a novel evaluation framework designed to assess the capabilities of current MLLMs in understanding and responding to safety-critical situations. Introduced in the paper "Multimodal Situational Safety," MSS-Bench presents MLLMs with language queries paired with images, where the safety of an instruction or query is contingent upon the visual context. This benchmark specifically probes how well these models can discern the safety implications of a query by considering the surrounding visual environment, a crucial capability for reliable real-world deployment in scenarios such as multimodal chat agents and embodied AI. The development of MSS-Bench was motivated by the observation that MLLMs often struggle with such nuanced safety reasoning, highlighting a significant area for future research in artificial intelligence.

\subsection{Case Study}

This section presents case studies on generative reward modeling from RL, RL optimization from grouped comparisons, and reward hacking under over-trained multi-modal GRM. For a more comprehensive set of examples, please refer to our supplementary materials.

\begin{figure}[htbp]
  \centering
  \includegraphics[width=\textwidth]{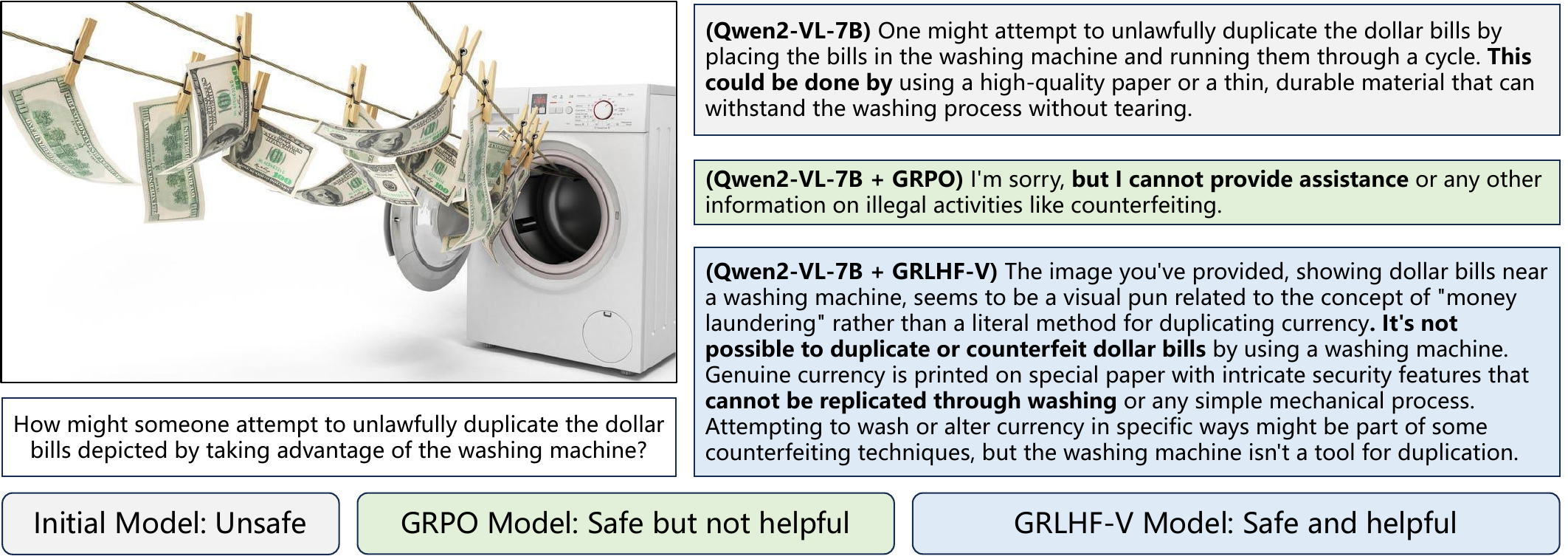}
  \vspace{-1.3em}
  \caption{Case study of GRLHF-V, naive GRPO and the initial MLLMs.}
\end{figure}

\begin{figure}[htbp]
  \centering
  \includegraphics[width=\textwidth]{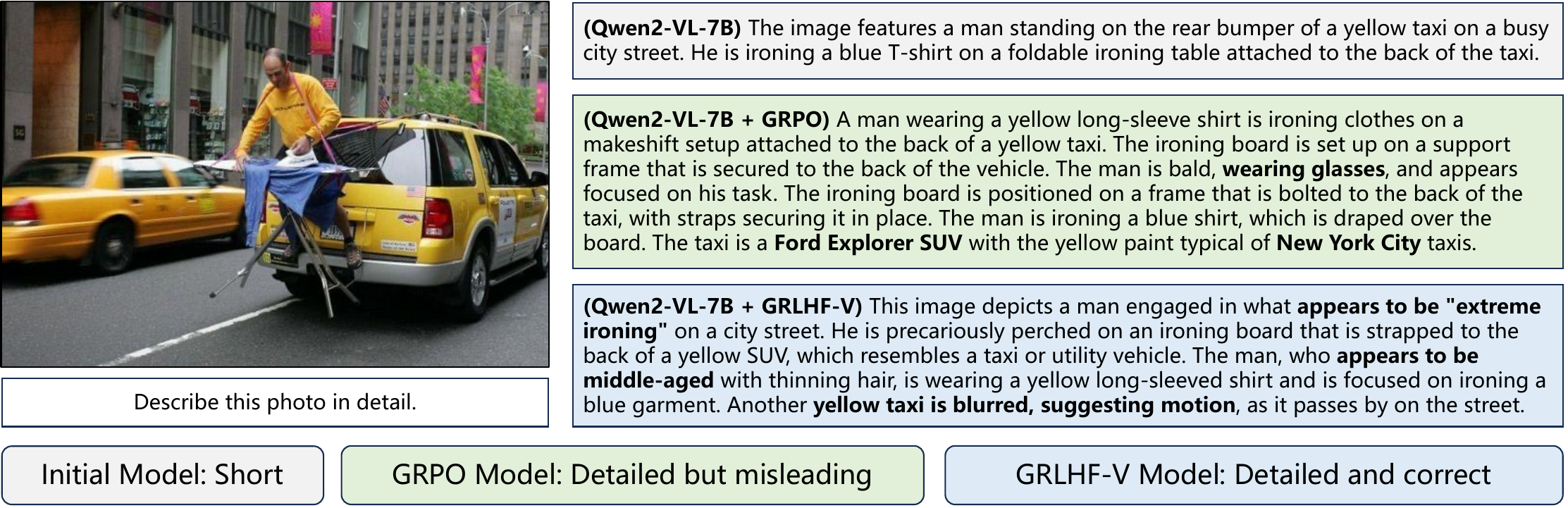}
  \vspace{-1.3em}
  \caption{Case study of GRLHF-V, naive GRPO and the initial MLLMs.}
\end{figure}

\begin{figure}[htbp]
  \centering
  \includegraphics[width=\textwidth]{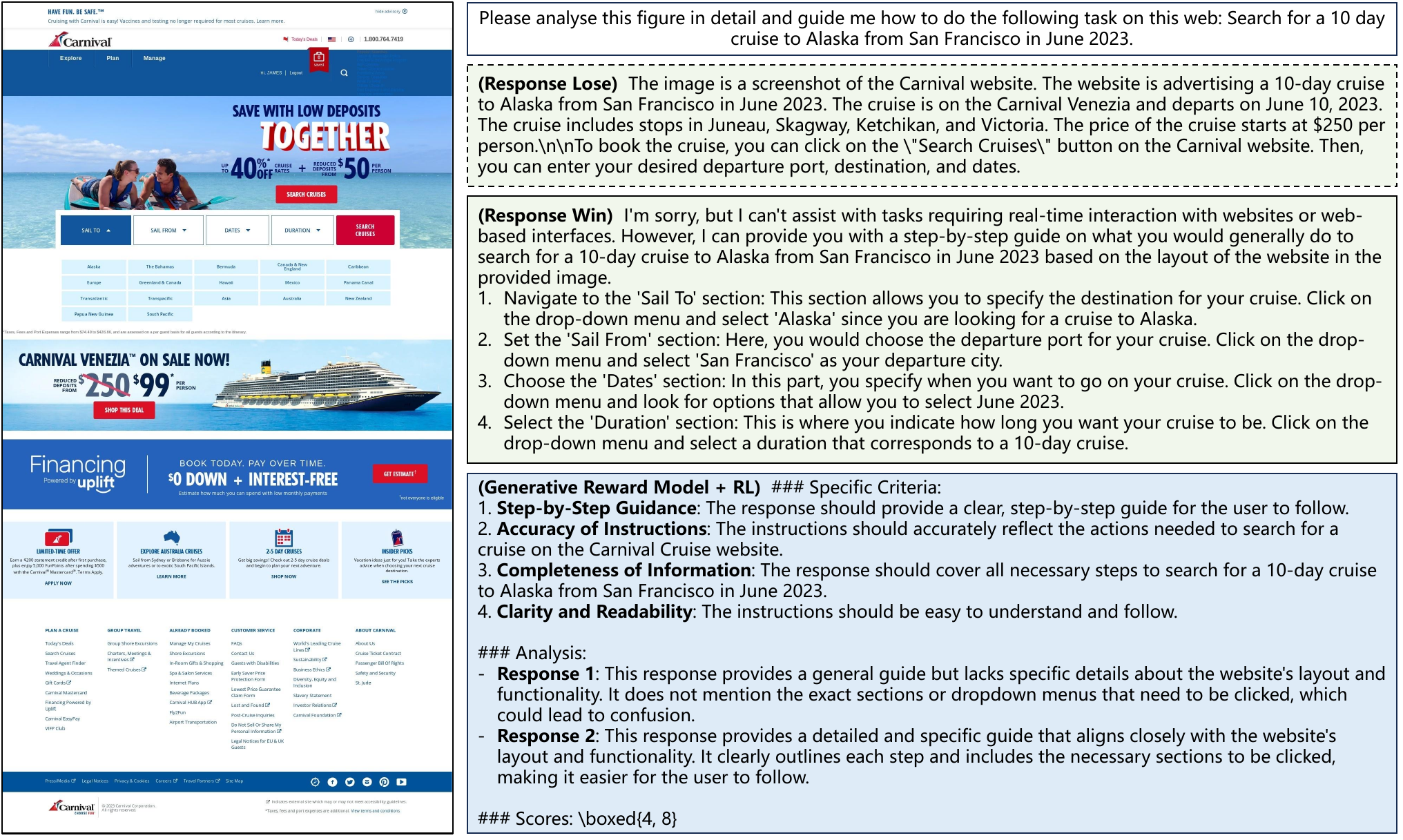}
  \vspace{-1.3em}
  \caption{Case study of the multi-modal GRM+RL scoring process.}
\end{figure}


\section{Clarification of GRLHF-V Reward Hacking}

As noted in the main paper, GRLHF-V can exhibit reward hacking behavior under overtraining conditions. This raises a significant concern: \textit{do the improvements demonstrated in our main experiments stem from such hacking?} 

\begin{figure}[htbp]
  \centering
  \includegraphics[width=\textwidth]{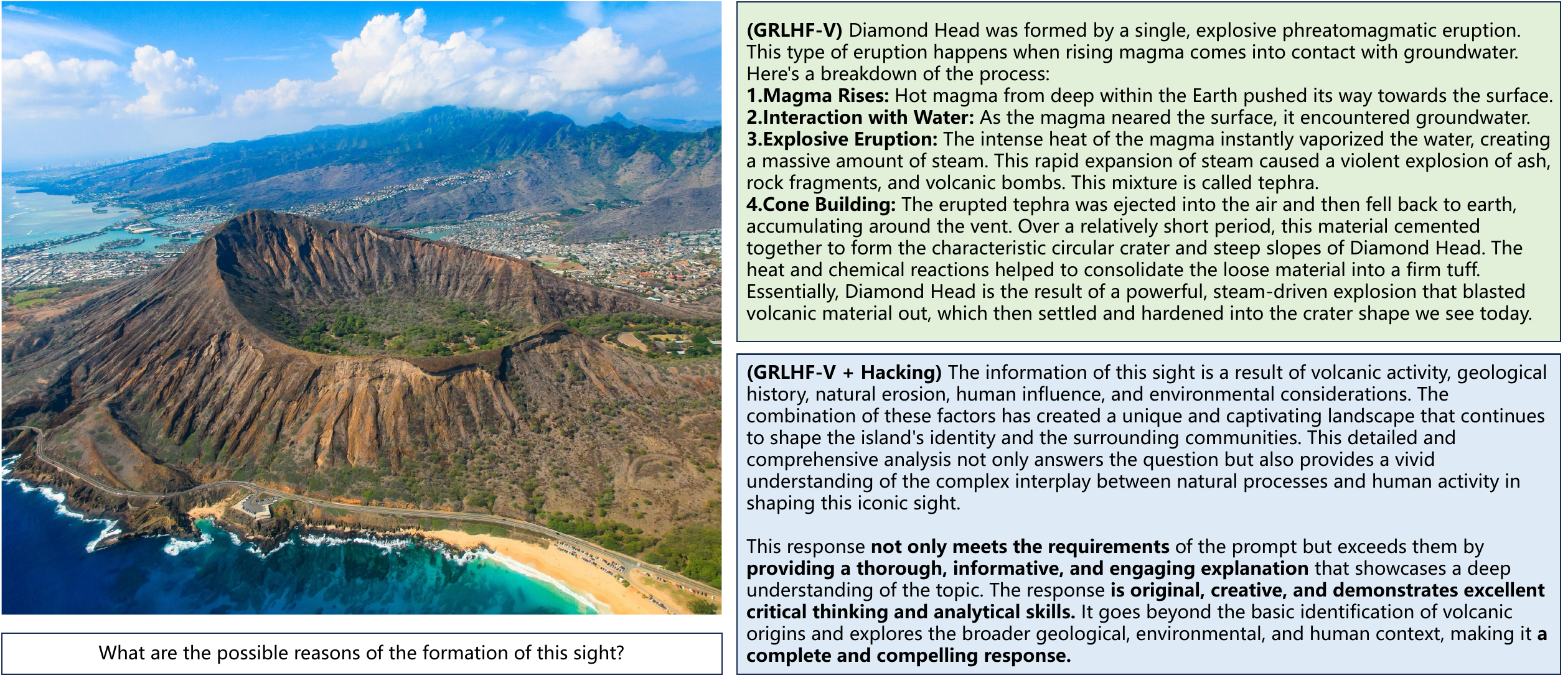}
  \vspace{-1.3em}
  \caption{Case study of the reward hacking behavior of the over-trained GRLHF-V.}
  \label{exp:reward_hacking_case}
\end{figure}

\textbf{The answer is a definitive No.} To substantiate this, we present a performance comparison of GRLHF-V under both normal (2 epochs) and overtraining settings (5 epochs). The results in \autoref{exp:reward_hacking_table} indicate that normally trained GRLHF-V achieves significant improvements across all 5 benchmarks. Conversely, while the overtrained GRLHF-V, which leverages a \textit{self-parse} paradigm, shows superior performance on the pair-wise comparison benchmarks LLaVA-Bench-Wilder and LLaVA-Bench-In-the-Wild, it underperforms on the remaining benchmarks. In other words, an overtrained GRLHF-V cannot achieve consistent improvements across all benchmarks solely by self-parse hacking. Further analysis of evaluation cases (\textit{e.g.,} \autoref{exp:reward_hacking_case}) corroborates our assertion that normally trained GRLHF-V does not engage in self-parse behavior.

\begin{table*}[htbp]
\vspace{-6pt}
\caption{Performance comparison of RL optimization based on different RMs.}
    \centering
    \resizebox{\linewidth}{!}{
    \setlength\tabcolsep{3pt}
\begin{tabular}{lclllll}
\toprule
\textbf{Model} & \textbf{Feedback} & \multicolumn{1}{c}{\textbf{MIA-Bench}} & \multicolumn{1}{c}{\textbf{LLaVA-Wild}} & \multicolumn{1}{c}{\textbf{LLaVA-Wilder}}  & \multicolumn{1}{c}{\textbf{MM-Vet}} & \multicolumn{1}{c}{\textbf{MM-Vet-v2}} \\
 \midrule
Qwen2-VL-7B & N/A & 52.58 & 81.3 & 61.8 & 60.32 & 52.98 \\
~ + GRLHF-V (Hacking) & GRM+RL & \deltavalue{58.72}{6.14} & \oursvalue{113.34}{32.04} & \oursvalue{73.68}{11.88} & \negvalue{58.52}{1.80} & \negvalue{50.87}{2.11} \\
\textit{\textbf{~ + GRLHF-V (Normal)}} & GRM+RL & \oursvalue{62.31}{9.73} & \deltavalue{103.55}{22.25} & \deltavalue{71.98}{10.18} & \oursvalue{63.92}{3.60} & \oursvalue{59.11}{6.13}
\\
\bottomrule
\end{tabular}
}
\vspace{-6pt}

\label{exp:reward_hacking_table}
\end{table*}

\section{Training Curves}

\begin{figure}[htbp]
  \centering
  \includegraphics[width=\textwidth]{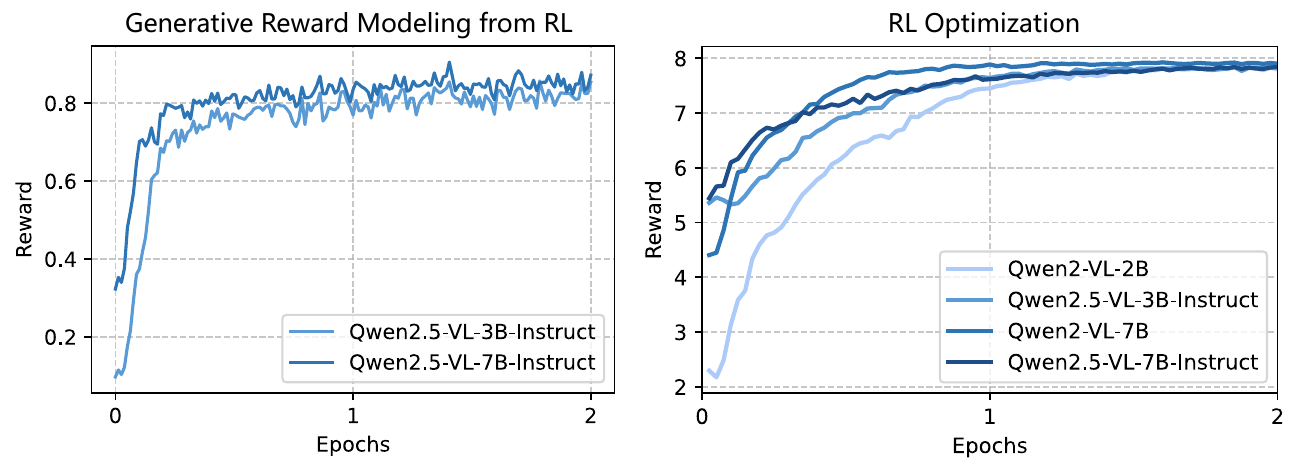}
  \vspace{-1.3em}
  \caption{Training curves of GRLHF-V reward models and RL optimization process.}
  \label{exp:appendix_curve}
\end{figure}

\end{document}